\documentclass{svproc}
\usepackage{url}

\usepackage[british]{babel}

\usepackage{xcolor}

\usepackage{graphicx}

\usepackage{multirow}

\usepackage{tabularx}

\usepackage{cite}

\usepackage{hyperref}

\usepackage{comment}
\begin{document}
\mainmatter              
\title{Locosim: an Open-Source Cross-Platform Robotics Framework}
\titlerunning{Locosim}  
%
\author{Michele Focchi\inst{1,}\inst{2,}\inst{*} \and Francesco Roscia\inst{1,}\inst{3,}\inst{*} \and Claudio Semini\inst{1}}
\authorrunning{Focchi, Roscia, Semini} 
%
\tocauthor{Michele Focchi, Francesco Roscia, Claudio Semini}
\institute{Dynamic Legged Systems (DLS), Istituto Italiano di Tecnologia (IIT), Genoa, Italy\\
\and
Department of Information Engineering and Computer Science (DISI), University of Trento, Trento, Italy\\
\and
Department of Informatics, Bioengineering, Robotics and Systems Engineering (DIBRIS), University of Genoa, Genova, Italy \\
\email{\{michele.focchi\}, \{francesco.roscia\}, \{claudio.semini\}@iit.it}\\
{\small $^*\,$The authors equally contributed to this paper}
}

\maketitle              
\begin{abstract}
The architecture of a robotics software framework tremendously influences the effort and time it takes for end users to test new concepts in a simulation environment and to control real hardware. 
Many years of activity in the field allowed us to sort out crucial requirements for a framework tailored for robotics: modularity and extensibility, source code reusability, feature richness, and user-friendliness.  
We implemented these requirements and collected best practices in Locosim, a cross-platform framework for simulation and real hardware. 
In this paper, we describe the architecture of Locosim and illustrate some use cases that show its potential.
\keywords{Computer Architecture for Robotics;
Software Tools for Robot Programming;
Software-Hardware Integration for Robot Systems}
\end{abstract}
\section{Introduction}
%
Writing software for robotic platforms can be arduous, time-consuming, and error-prone. 
In recent years, the number of research groups working in robotics has grown exponentially, each group having platforms with peculiar characteristics. 
The choice of morphology, actuation systems, and sensing technology is virtually unlimited, and code reuse is fundamental to getting new robots up and running in the shortest possible time. 
In addition, it is pervasive for researchers willing to test new ideas in simulation without wasting time in coding, for instance, using high-level languages for rapid code prototyping. 
To pursue these goals, in the past years several robotics frameworks have been designed for teaching or for controlling specific platforms, e.g., OpenRAVE \cite{openrave}, Drake \cite{drake} and SL \cite{Schaal2009}. 

To avoid roboticists \textit{reinventing the wheel} whenever they buy or build a new robot, 
we present our framework Locosim\footnote[1]{Locosim  can be downloaded from \url{www.github.com/mfocchi/locosim}.}. 
Locosim is designed with the primary goal of being platform-independent, 
dramatically simplifying the task of interfacing a robot with planners and controllers. 
Locosim consists of a ROS control node \cite{quigley2009ros} (the \textit{low-level} controller), written in C++, 
that interfaces a custom Python ROS node (the \textit{high-level} planner/controller) 
with either a Gazebo simulator \cite{koenig2004design} or  the real hardware. 
The planner relies on Pinocchio \cite{carpentier2019pinocchio} for computing the robot's kinematics and dynamics and closes the control loop at a user-defined frequency. 
\begin{figure}[t!]
    \centering
    \includegraphics[width=\textwidth]{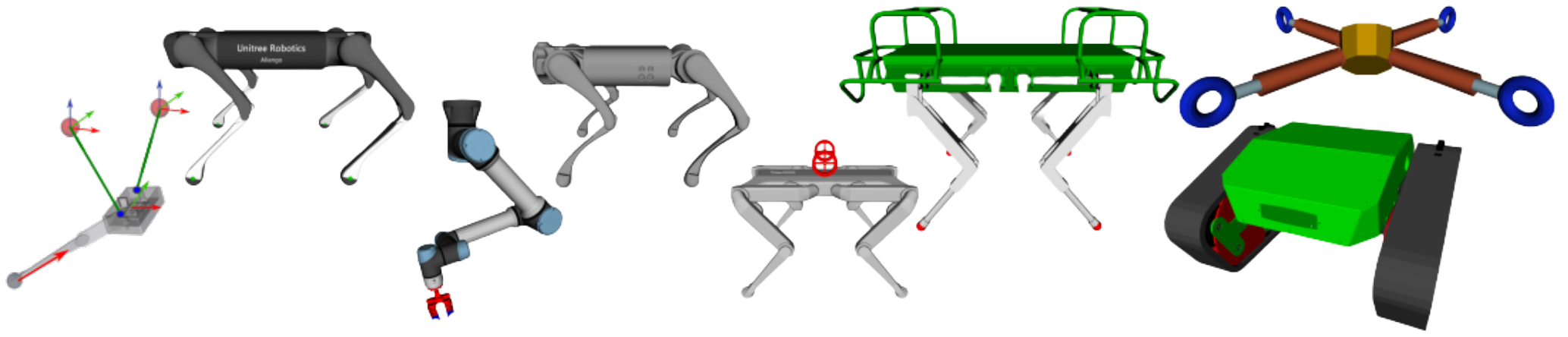}
    \caption{Examples of robots already included in Locosim 
    (from left to right, top to bottom): 
    Aliengo \cite{aliengo}, Go1 \cite{go1}, HyQ \cite{semini2011design}, Starbot, CLIO \cite{focchi2022clio}, UR5 \cite{ur5} with Gripper, Solo with Flywheels \cite{roscia2023orientation}, Tractor (images are not in scale). }
    \label{fig:robots}
    
\vspace{-1.5em} \end{figure}
\subsection{Advantages of Locosim}
The benefits of the proposed framework are multiple.
\begin{itemize}
    \item Locosim is platform-independent. It supports a list of robots with different morphology (e.g., quadrupeds, arms, hybrid structures, see Fig. \ref{fig:robots}) and provides features for fast designing and adding new robots.
    \item Locosim implements functions needed for all robots. Once the robot description is provided, no effort is spent on libraries for kinematics, dynamics, logging, plotting, or visualization. These valuable tools ease the synthesis of a new planner/controller.
    \item Locosim is easy to learn. The end user invests little time in training and gets an open-source framework with all the benefits of Python and ROS.
    \item Locosim is modular. Because it heavily uses the inheritance paradigm, classes of increasing complexity can provide different features depending on the nature of the specific robotic application. For instance, the controller for fixed-base robotic arms with a grasping tool can be reused for a four-legged robot with flywheels since it is built upon the same base class.
    \item Locosim is extensible. Our framework is modifiable and the end-user can add any supplementary functionality.
    \item Locosim is easy to install. It can be used  either inside a virtual machine, a docker container, or natively by manually installing dependencies.
\end{itemize}
\subsection{Outline}
The remainder of this paper is organized as follows.
In Section \ref{sec:key_aspects} we highlight the critical requirements of a cross-platform robotics framework. In Section \ref{sec:architecture} we detail structure and features of Locosim. 
In Section \ref{sec:use_cases} we discuss use-case examples of our framework, either with the real robot or with its simulated digital twin. Eventually, we condense the results and present future works in Section \ref{sec:conclusions}.
\begin{figure}[t]
    \centering
    \includegraphics[width=0.8\textwidth]{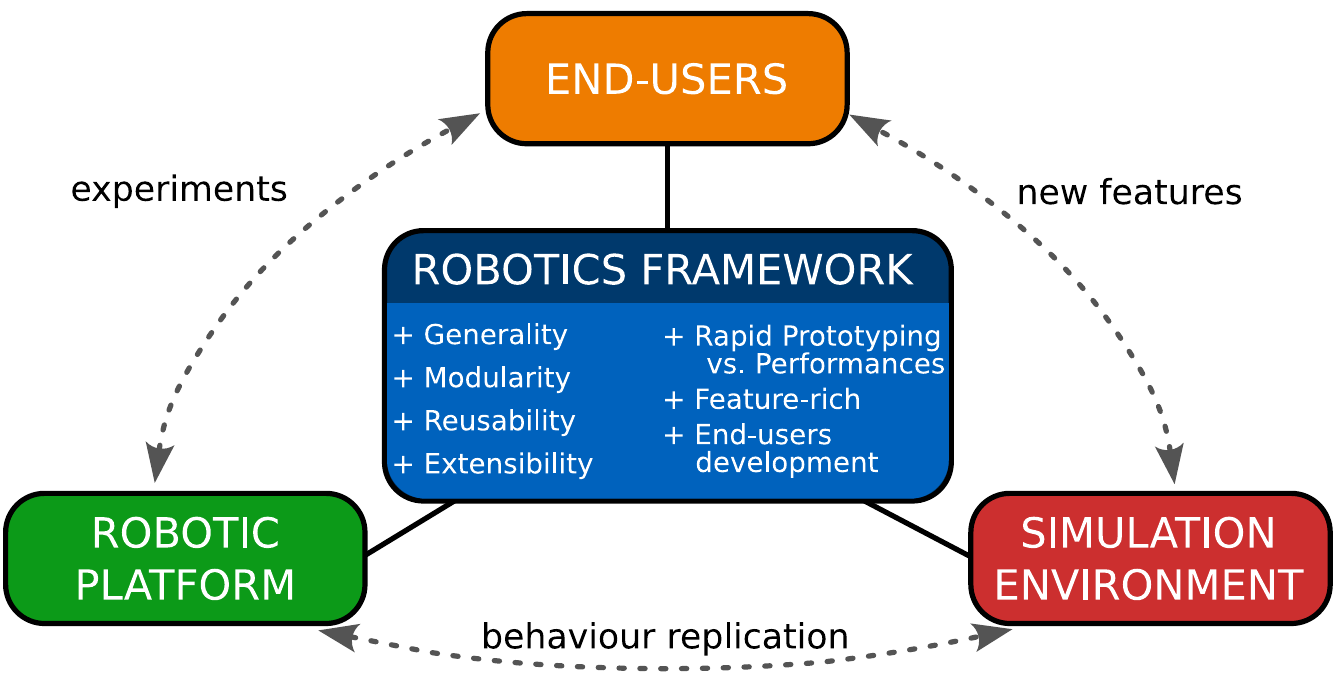}
    \caption{End-users, robotic platform and simulation environment make a triad only if an effective robotics framework can join them.}
    \label{fig:triangle}
    
\vspace{-1.5em} \end{figure}
\section{Key aspects of a robotics framework}
\label{sec:key_aspects}
In the most general sense, a \textit{robotics framework} is a software 
architecture of programs and data that adhere to well-defined rules for operating robots.
\textit{End-users} are people who will ultimately use the robotics framework and potentially bring some modifications.
The robotics framework is the center of a triangle having at its vertices 
the end-user, the robotic platform and the simulation environment (see Fig. \ref{fig:triangle}). 
The simulation must replicate the behaviour of the robot with a sufficient degree of accuracy. 
The end-user must be able to test new features and ideas in the simulation environment before running them on the robot platform.
A robotics framework should provide the link among these three. 
In this context, we identify a list of crucial requirements that 
a robotics framework must possess: generality, modularity, reusability, extensibility, rapid prototyping vs. performances, feature-rich, and end-users development.
\subsubsection{Generality.} 
It is essential to release software free from unnatural restrictions and limitations. 
An end-user may require to design software for controlling a single robot, for multi-robot cooperation, for swarm coordination, or for reinforcement learning, which requires an abundant number of robots in the training phase. It should be possible to model any kinematic structure (floating base/fixed base robot, kinematic loops, etc.). 

\subsubsection{Modularity.} 
A robotics framework should provide separate building blocks of functionalities according to the specific needs of the robot or the application. 
These building blocks must be replaceable and separated by clear interfaces.
The end-user may want to only visualize a specific robot in a particular joint configuration, move the joints interactively, or plan motions and understand the effects of external forces. 
Each of these functions must be equipped with tools for debugging and testing, e.g., unit tests.
Replacing each module with improved or updated versions with additional functionalities should be easy.

\subsubsection{Reusability.} 
Pieces of source code of a program should be reused by reassembling them in more complex programs to provide a desired set of  functionalities, with minor or no modifications.
From this perspective, parametrization is a simple but effective method, e.g., the end-user should be able to use the same framework with different robots changing only a single parameter. 
In the same way, \textit{digital twins} \cite{ramasubramanian2022digital} can be realized by varying the status of a flag that selects between the real hardware and the simulation environment. 
This avoids writing different codes for the simulator and the real robot. 

\subsubsection{Extensibility.} A robotics framework must be designed with the foresight 
to support the addition and evolution of hardware and software that may not exist at implementation time. 
This property can be achieved by providing a general set of application programming interfaces (APIs).
Concepts typical of Object-Oriented Programming, such as inheritance and polymorphism, play a crucial role in extensibility. 

\subsubsection{Rapid Prototyping vs. Performances.} 
A framework should allow for fast code prototyping of researchers' ideas. 
More specialized controllers/planners are built from simpler ones in the form of 
recursive inheritance (\textit{matryoshka principle}). 
In this way end–users have unit tests at different levels of complexity. 
With fast code prototyping, end-users can quickly write software without syntax and logic errors.
However, they do not have any assurance about the performance: such code is just good enough for the case of study in a simulation environment. 
Stringent requirements appear when executing codes on real robots, e.g., short computation time and limited memory usage. 
Thus, the framework must expose functionalities that can deal with performance.

\subsubsection{Feature-rich.} Most of the end-users need a sequence of functionalities when working with robots.
These include but are not limited to the computation of kinematics and dynamics, logging, plotting, and visualization. 
A robotics framework should provide them, and they must be easily accessible.

\subsubsection{End-users Development.} 
Besides its implementation details, a robotics framework should provide methods, techniques and tools that allow end-users to create, modify, or extend the software \cite{Lieberman2006} in an easy way.
It should run on widely used Operating Systems and employ renowned high-level programming languages that facilitate software integration.
Clear documentation for installation and usage must be provided, and modules should have self-explanatory names and attributes. 
\section{Locosim Description}
\label{sec:architecture}
Locosim was born as a didactic framework to simulate both fixed- and floating-base robots. 
Quickly it evolved to be a framework for researchers that want to program newly purchased robots in a short time. 
Locosim runs on machines with Ubuntu as Operating System, and it employs ROS as middleware. Within Locosim, end-users can write robot controllers/planners in Python. 
\subsection{Architecture}
Locosim consists of four components: Robot Descriptions, Robot Hardware Interfaces, ROS Impedance Controller, and Robot Control. We illustrate each component in the following\footnote{Some of the functions in Locosim components, which are quite established for the robotics community, are named after \cite{Schaal2009}.}. 

\subsubsection{Robot Descriptions.} 
The Robot Descriptions component contains dedicated packages for the characterization of each robot. For instance, the package that  contains files to describe the robot \texttt{myrobot}, a generic mobile robot platform, is named \texttt{myrobot\_description}.
With the main focus on fast prototyping and human readability, the robot description is written in Xacro (XML-based scripting language) that avoids code replication through macros, conditional statements and parameters.
It can import descriptions of some parts of the robot
from \texttt{urdfs} sub-folder or meshes describing the geometry of rigid bodies from the \texttt{meshes} sub-folder. 
At run time, the Xacro file is converted into URDF, allowing the end-user to 
change some parameters. The \texttt{gazebo.urdf.xacro} launches 
\texttt{ros\_control} package and the \texttt{gazebo\_ros\_p3d} 
plugin, which publishes the pose of the robot trunk in 
the topic \texttt{/ground\_truth} (needed only for floating-base robots).
The robot description directory must contain the files \texttt{upload.launch} and \texttt{rviz.launch}. The former processes the Xacro generating the URDF and loads into the parameter server, and the latter allows to visualize the robot and interact with it by providing the \texttt{conf.rviz} file. 
This is the configuration for the ROS visualizer RViz, which can be different for every robot. 
Additionally, the file \texttt{ros\_impedance\_controller.yaml} must be provided for each robot: it contains the sequence of joint names, joint PID gains and the home configuration. 
The Python script \texttt{go0} will be used during the simulation startup to drive the robot to the home configuration.

\subsubsection{Robot Hardware Interfaces.} 
This folder contains drivers for the real hardware platforms supported by Locosim. 
They implement the 
interface that bridges the communication between the controller and the real robot, abstracting the specificity of  each robot and exposing the same interface (e.g., \texttt{EffortInterface}). 
For instance, the UR5 robot through its driver provides three possible ROS hardware interfaces: Effort, Position and Velocity, hiding the details of the respective underlying low-level controllers.
\begin{table}
    \caption{Main attributes and methods of the \texttt{BaseControllerFixed} (BCF) and of the \texttt{BaseController} (BC) classes. All vectors (unless specified) are expressed in world frame.  For methods with the same name, the derived class loads the method of the parent class and adds additional elements specific to that class.}
\begin{tabularx}{\textwidth}{llp{0.5em} p{17em}p{0.5em}l}
\hline
& Name & & Meaning & & Class\\
 \hline
 \hline
\parbox[t]{1.5em}{\multirow{12}{*}{\rotatebox[origin=c]{90}{Attributes}}} 
& \texttt{q}, \texttt{q\_des} & & actual / desired joint positions & & BCF, BC\\
& \texttt{qd}, \texttt{qd\_des} & & actual /desired joint velocities & & BCF, BC\\ 
& \texttt{tau}, \texttt{tau\_ffwd} & & actual / feed-forward joint torques & & BCF, BC\\
& \texttt{x\_ee} & & position of the end-effector expressed in base frame & & BCF\\
& \texttt{contactForceW} & & contact force at the end-effector  & & BCF \\
& \texttt{contactMomentW} & & contact moment at the end-effector  & & BCF \\
& \texttt{basePoseW} & & base position and orientation in Euler angles & & BC \\
& \texttt{baseTwistW} & & base linear and angular velocity & & BC\\
& \texttt{b\_R\_w} & & orientation of the base link  & & BC\\
& \texttt{contactsW} & & position of the contacts   & & BC\\
& \texttt{grForcesW} & & ground reaction forces on contacts   & & BC\\
\hline
\parbox[t]{1.5em}{\multirow{27}{*}{\rotatebox[origin=c]{90}{Methods}}} 
& \texttt{loadModelAndPublishers()} & & creates the object \texttt{robot}  (Pinocchio wrapper) and loads publishers for visual features (\texttt{ros\_pub}), joint commands and declares subscriber to \texttt{/ground\_truth} and  \texttt{/joint\_states} & & BCF, BC\\
& \texttt{startFramework()} & & launch \texttt{ros\_impedance\_controller} & & BCF, BC\\
& \texttt{send\_des\_jstate()} & & publishes \texttt{/command} topic with set-points for joint positions, velocities and feed-forward torques & & BCF, BC\\
& \texttt{startupProcedure()} & & initializes PID gains & & BCF, BC\\
& \texttt{initVars()} & & initializes class attributes & & BCF, BC\\
& \texttt{logData()} & & fill in the variables \texttt{x\_log} with the content of  \texttt{x} for plotting purposes, it needs to be called at every loop & & BCF, BC\\
& \texttt{receive\_jstate()} & & callback associated to the subscriber \texttt{/joint\_states}, fills in actual joint positions, velocities and torques & & BCF, BC\\
& \texttt{receive\_pose()} & & callback associated to the subscriber \texttt{/ground\_truth}, fills in the actual base pose and twist, and publishes the fixed transform between world and base & & BC\\
& \texttt{updateKinematics()} & & from \texttt{q}, \texttt{qd}, \texttt{basePoseW}, \texttt{baseTwistW}, computes position and velocity of the robot's center of mass, the contacts position, the associated Jacobians, the ground reaction forces, the centroidal inertia, the mass matrix and the non-linear effects & & BC\\
\hline
\end{tabularx}
\label{tab:attributes_methods}
\end{table}
\begin{figure}[t!]
    \centering
    \includegraphics[width=\textwidth]{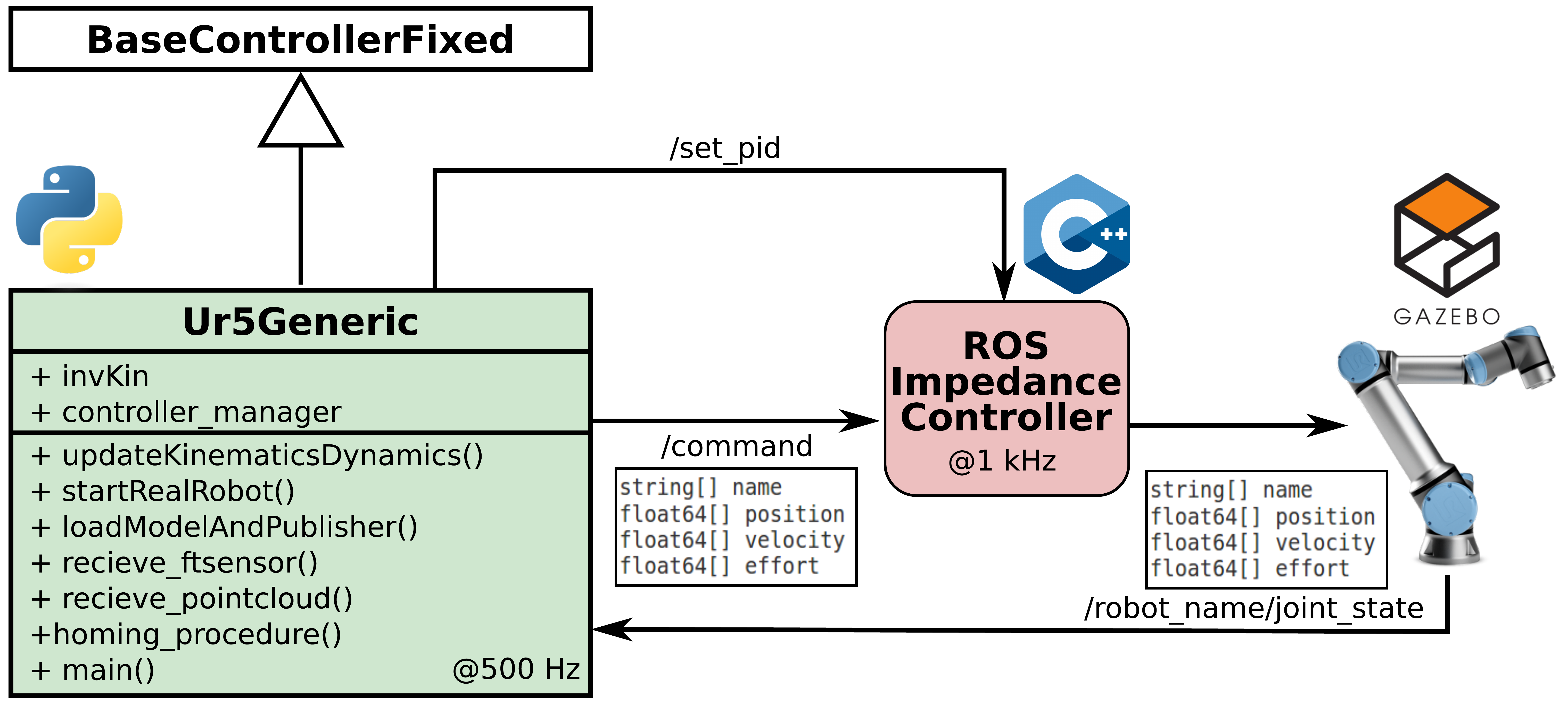}
    \caption{Schematic representation of a typical use-case of Locosim. The end-user wants to simulate the UR5 robot arm. An instance of \texttt{Ur5Generic}, which is a derived class of \texttt{BaseControllerFixed}, sends the command to the robot though \texttt{ros\_impedance\_controller} and it receives back the actual state.
    \texttt{Ur5Generic} implements features to manage the real robot and the gripper, perform a homing procedure at startup and a class for inverse kinematics. 
    } 
    \label{fig:uml}
    
\vspace{-1.5em} \end{figure}
\subsubsection{ROS Impedance Controller.} 
ROS Impedance Controller is a ROS package written in C++ 
that implements the \textit{low-level} joint controller, 
in the form of PID with feedforward effort (force or torque).
The \texttt{/joint\_state\_publisher} publishes the 
actual position, velocity and effort for each robot joint. 
By default, the loop frequency is set to 1 kHz. 
This and other parameters can be regulated in the launch file called \texttt{ros\_impedance\_controller.launch}. 
Robots with specific needs can be dealt with by specifying a custom launch file. This is the case of the CLIO climbing robot that requires the model of the mountain to which it is attached.
The \texttt{robot\_name} parameter is used to load  the correct robot description. 
If the \texttt{real\_robot} flag is set to true, the robot hardware interface is initialized; otherwise, the Gazebo simulation starts running first the \texttt{go0} script from the robot description. 
In any case, Rviz will be opened with the robot's specific configuration file \texttt{conf.rviz}.
The end-user can manually change the location where the 
robot is spawned in Rviz with the \texttt{spawn} parameters. 
Physics parameters for the simulator are stored in the 
sub-folder \texttt{worlds}.

\subsubsection{Robot Control.} 
From the end-user perspective, this is the most crucial component.
It embraces classes and methods for computation of the robot's kinematics and dynamics, logging and plotting time series variables, and real-time visualization on Rviz. 
Within this component, \textit{high-level} planning/control strategies are implemented.
The codes of the component are entirely written in Python and have a few dependencies: above all, NumPy \cite{harris2020array} and Pinocchio \cite{carpentier2019pinocchio}. 
The former offers tools for manipulating multidimensional arrays; the latter implements functions for the robot's kinematics and dynamics.
Pinocchio can be an essential instrument for researchers because of its  efficient  computations.
Nevertheless, it can be  time-consuming and cumbersome to understand for newcomers. 
To facilitate the employment, we developed a \texttt{custom\_robot\_wrapper} for building a \texttt{robot} object, computing robot mass, center of mass, Jacobians, centroidal inertia, and so on with easy-to-understand interfaces.\\
For the end-user, the starting point for building up a robot planner is the class \texttt{BaseControllerFixed}, suitable for fixed-base robots, and the derived class \texttt{BaseController}, which handles floating-base ones.
In Table \ref{tab:attributes_methods}, we report a list of the main attributes and methods of \texttt{BaseController} and of its parent \texttt{BaseControllerFixed}. 
For having a more complex and specific controller, the end-user can create its own class, inheriting from one of the previous two and adding additional functionalities.
E.g., \texttt{QuadrupedController} inherits from \texttt{BaseController}, and it is specific for quadruped robots.
\texttt{Ur5Generic} adds to \texttt{BaseControllerFixed} the features of controlling the real robot, a gripper and a camera attached (or not) to the robotic arm UR5
(see Fig. \ref{fig:uml}). 
The controller class is initialized with the string \texttt{robot\_name}, e.g., we write \texttt{BaseController(myrobot)} for the controller of \texttt{myrobot}. 
The end-user must pay particular attention to this string because it creates the link with the robot description and the robot hardware interface if needed. 
The \texttt{robot\_name} is used for accessing the dictionary \texttt{robot\_param} too. 
Among the other parameters of the dictionary, the flag \texttt{real\_robot} permits using the same code for both the real (if set to true) and simulated (false) robot, resulting in the digital twin concept. 
The \texttt{BaseControllerFixed} class contains a \texttt{ControllerManager} to seamlessly swap between the control modes and controller types if the real hardware supports more than one. For instance, UR5 has two control modes (point, trajectory) and two controller types (position, torque), whereas Go1 supports a single low-level torque controller.
Additionally, the \texttt{GripperManager} class manages the gripper in different ways for the simulation (i.e., adding additional finger joints) and for the real robot (on-off opening/closing service call to the UR5 driver as specified by the manufacturer), hiding this complexity to the end-user. 
The method \texttt{startFramework()} permits to launch the simulation or the driver, executing \texttt{ros\_impedance\_controller.launch}. 
It takes as input the list \texttt{additional\_args} to propagate supplementary arguments, dealing  with robot and task specificity.
In the \texttt{components} folder there are additional classes for inverse kinematics, whole-body control, leg odometry, IMU utility, filters and more. 
Finally, the folder \texttt{lab\_exercises} contains an ample list of scripts, with didactic exercises of incremental complexity, to learn Locosim and the main robotics concepts.
\subsection{Analysis of Design Choices}
To fulfill the requirements stated in Section 
\ref{sec:key_aspects}, we made a number of choices. We want to focus on why we selected ROS as middleware, Python as (preferred) programming language, and Pinocchio as computational tool for the robot's kinematics and dynamics. 

\subsubsection{Why ROS.} 
The ROS community is spread worldwide. 
Over the last decade, the community produced hundreds or thousands of open-source tools and software: from device drivers and interface packages of various sensors and actuators to tools for debugging, visualization, and off-the-shelf  algorithms for multiple purposes. 
With the notations of nodes, publishers, and subscribers, 
ROS quickly solves the arduous problem of resource handling between many processes. 
ROS is reliable: the system can still work if one node crashes. 
On the other hand, learning ROS is not an effortless task for newcomers. 
Moreover, modeling robots with URDF can take lots of time, as well as starting simulations \cite{joseph2018mastering}. 
Locosim relieves end-users from these complications by adopting 
a \textit{common skeleton} infrastructure for the robot description and for the high-level planner/controller.

\subsubsection{Why Python.} 
Among the general-purpose programming languages, Python is one of the most used. 
It is object-oriented and relatively straightforward to learn and use compared to other languages. 
The availability of open-source libraries and packages is virtually endless.
These reasons make Python perfect for fast prototyping software. 
Being an interpreted language, Python may suffer in terms of 
computation time and memory allocation, colliding with the real-time requirements of the real hardware. 
In these cases, when testing the code in simulation, the end-user 
may consider using profiling tools to look for the most demanding parts of the code.
Before executing the software on the real robot, these parts 
can be optimized within Python or translated into C++ code, providing Python bindings. 
For the same performance reasons, the most 
critical part of the framework, 
the low-level controller, is directly implemented in C++. 

\subsubsection{Why Pinocchio.} Pinocchio \cite{carpentier2019pinocchio} is one of the most efficient tools for computing poly-articulated bodies' kinematics and dynamics.
Differently from other libraries in which a meta-program generates some source code for a single specific robot model given in input, as in the case of RobCoGen \cite{frigerio2016robcogen}, Pinocchio is a dynamic library that loads at runtime any robot model. This characteristic makes it suitable for a cross-platform framework.
It implements the state-of-the-art Rigid Body Algorithms based on revisited Roy Featherstone’s algorithms \cite{featherstone2014rigid}.
Pinocchio is open-source, mostly written in C++ with Python bindings and several projects rely on it. 
Pinocchio also provides derivatives of the kinematics and dynamics, which are essential to in gradient-based optimization.

\section{Use Cases}
\label{sec:use_cases}
We want to emphasize the valuable features of Locosim with practical 
use cases\footnote[2]{A video showing the above-mentioned and other use cases can be found here:
\begin{sloppypar} 
\noindent \url{https://youtu.be/ZwV1LEqK-LU}
\end{sloppypar}}.

\subsection{Visualize a Robot: kinematics check}
As first use case, we illustrate the procedure to visualize 
the quadruped robot Aliengo (see Fig. \ref{fig:robots}) in RViz and how to manually interact with its joints.
This is a debugging tool which is crucial during the
design process of a new robot, because it allows to test the kinematics without added complexity. 
In the Robot Description package, we create a folder named \texttt{aliengo\_description}. 
In the  \texttt{robots} folder we add the XML file for describing the robot's kinematic and dynamic structure. 
We make use of the flexibility of the open source Xacro language to simplify writing process: we include files that describe a leg, the transmission model, and the meshes for each of the bodies. 
We create the \texttt{launch} folder containing the files \texttt{upload.launch} and \texttt{rviz.launch}.
Launching \texttt{rviz.launch} from command line, the end-user can visualize the robot in RViz 
and manually move the joints by dragging the sliders of the \texttt{joint\_state\_publisher\_gui}.
The \texttt{conf.rviz} file helps the end-user to restore previous sessions in RViz.
With this simple use case we can effectively understand the importance of the key aspects formalized so far. 
Being extensible, Locosim allows for the introduction of any new robots, reusing parts of codes already present. 

%
\begin{figure}[t!]
    \centering
    \includegraphics[width=.8\textwidth]{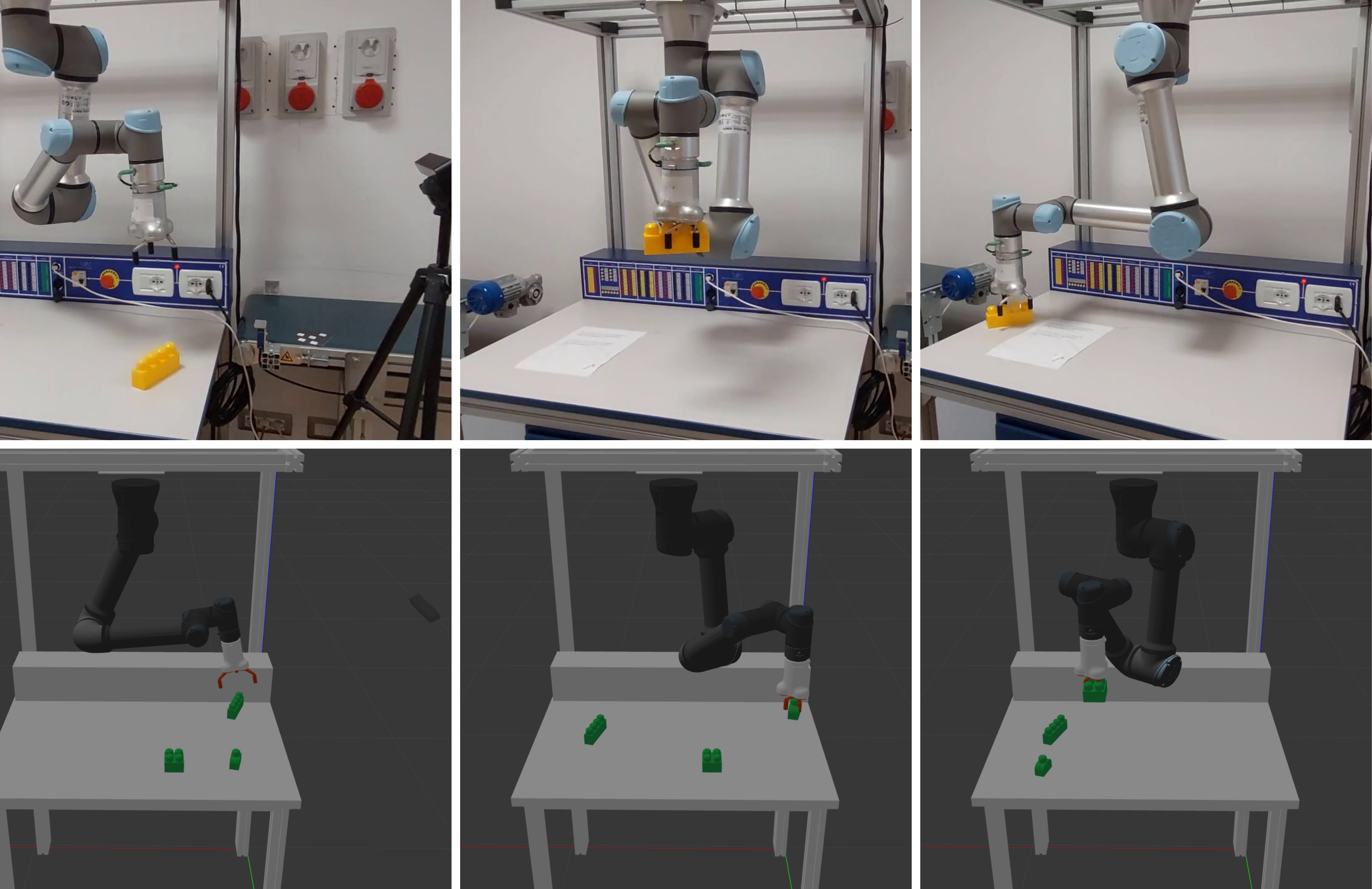} 
    \caption{Execution of the pick-and-place task with the anthropomorphic arm  UR5. The end-user can drive the real hardware (setting \texttt{real\_robot} to true) or perform a simulation (\texttt{real\_robot} to false).} 
    \label{fig:ur5}
    
\vspace{-1.5em} \end{figure}
\subsection{Simulation and Real Robot}
As a second example, we present a pick-and-place task with the anthropomorphic arm UR5 (6 degrees of freedom, see Fig. \ref{fig:robots}).
The pipeline of planning and control starts with the launch file \texttt{ros\_impedance\_controller.launch}. 
This is common for all the robots: it loads the \texttt{robot\_description} parameter calling the \texttt{upload.launch} and it starts the Gazebo simulator or the 
robot driver according to the status of the \texttt{real\_robot} flag.
Additionally, it loads the controller parameters (e.g., PID gains), which are located in the \texttt{robot\_description} package of the robot. 
In the simulation case, the launch file spawns the robot 
at the desired location. 
In the real robot case, the robot driver is running (in a ROS node) on an onboard computer while the planner runs on an external computer. 
In both cases, the RViz GUI shows the robot configuration.
Another node running on the same computer reads from a fixed frame ZED2 camera and publishes a point cloud message of the environment on a ROS topic.
We extract the coordinates of the plastic bricks that are present in the workspace. 
With \texttt{Ur5Generic}, we plan trajectories for 
the  end-effector position and orientation to grasp and  relocate the bricks. 
We set an inverse kinematic problem to find a joint reference trajectory.
It is published in the \texttt{/ur5/command} topic, together with feed-forward torques for gravity compensation.
The low-level \texttt{ros\_impedance\_controller} 
provides feedback for tracking the joint 
references, based on the actual state in \texttt{/joint\_state}. 
On the real robot there is no torque control and only position set-points are provided to the \texttt{ur5/joint\_group\_pos\_controller/command} topic as requested by the robot driver provided by the manufacturer. 
All this is dealt with by the \texttt{controller\_manager} class, 
transparent to the end users.
The power of Locosim lies on the fact that it is possible
to use the same robot control code both to simulate the task and to execute it on the real robot, as reported in Fig. \ref{fig:ur5}.

\section{Conclusions}
\label{sec:conclusions}
Locosim is a platform-independent framework for working with robots, either in a simulation environment or with real hardware. 
Integrating features for computation of robots' kinematics and dynamics, 
logging, plotting, and visualization, Locosim is a considerable 
help for roboticists that needs a starting point for rapid code prototyping.  
If needed, performances can be ensured by implementing the critical parts of the software
in C++, providing Python bindings. 
We proved the usefulness and versatility of our framework with use cases. 
Future works include the following objectives: Locosim will be able to handle multiple platforms simultaneously to be used in the fields of swarm robotics and collaborative robotics. 
We want to provide support for ROS2 since ROS lacks relevant qualifications such as real-time, safety, certification, and security. 
Additionally, other simulator like Coppelia Sim or PyBullet will be added to Locosim.

\bibliographystyle{IEEEtran}
\bibliography{references/bibliography}

\begin{thebibliography}{10}
\providecommand{\url}[1]{#1}
\csname url@samestyle\endcsname
\providecommand{\newblock}{\relax}
\providecommand{\bibinfo}[2]{#2}
\providecommand{\BIBentrySTDinterwordspacing}{\spaceskip=0pt\relax}
\providecommand{\BIBentryALTinterwordstretchfactor}{4}
\providecommand{\BIBentryALTinterwordspacing}{\spaceskip=\fontdimen2\font plus
\BIBentryALTinterwordstretchfactor\fontdimen3\font minus
  \fontdimen4\font\relax}
\providecommand{\BIBforeignlanguage}[2]{{%
\expandafter\ifx\csname l@#1\endcsname\relax
\typeout{** WARNING: IEEEtran.bst: No hyphenation pattern has been}%
\typeout{** loaded for the language `#1'. Using the pattern for}%
\typeout{** the default language instead.}%
\else
\language=\csname l@#1\endcsname
\fi
#2}}
\providecommand{\BIBdecl}{\relax}
\BIBdecl

\bibitem{openrave}
\BIBentryALTinterwordspacing
``{OpenRAVE}.'' [Online]. Available: \url{https://github.com/rdiankov/openrave}
\BIBentrySTDinterwordspacing

\bibitem{drake}
\BIBentryALTinterwordspacing
``{Drake}.'' [Online]. Available:
  \url{https://github.com/RobotLocomotion/drake}
\BIBentrySTDinterwordspacing

\bibitem{Schaal2009}
\BIBentryALTinterwordspacing
S.~Schaal, ``The sl simulation and real-time control software package,'' Los
  Angeles, CA, Tech. Rep., 2009, clmc. [Online]. Available:
  \url{http://www-clmc.usc.edu/publications/S/schaal-TRSL.pdf}
\BIBentrySTDinterwordspacing

\bibitem{quigley2009ros}
M.~Quigley, K.~Conley, B.~Gerkey, J.~Faust, T.~Foote, J.~Leibs, R.~Wheeler,
  A.~Y. Ng \emph{et~al.}, ``Ros: an open-source robot operating system,'' in
  \emph{ICRA workshop on open source software}, vol.~3, no. 3.2.\hskip 1em plus
  0.5em minus 0.4em\relax Kobe, Japan, 2009, p.~5.

\bibitem{koenig2004design}
N.~Koenig and A.~Howard, ``Design and use paradigms for gazebo, an open-source
  multi-robot simulator,'' in \emph{2004 IEEE/RSJ International Conference on
  Intelligent Robots and Systems (IROS)(IEEE Cat. No. 04CH37566)},
  vol.~3.\hskip 1em plus 0.5em minus 0.4em\relax IEEE, 2004, pp. 2149--2154.

\bibitem{carpentier2019pinocchio}
J.~Carpentier, G.~Saurel, G.~Buondonno, J.~Mirabel, F.~Lamiraux, O.~Stasse, and
  N.~Mansard, ``The {P}inocchio {C}++ library: A fast and flexible
  implementation of rigid body dynamics algorithms and their analytical
  derivatives,'' in \emph{2019 IEEE/SICE International Symposium on System
  Integration (SII)}.\hskip 1em plus 0.5em minus 0.4em\relax IEEE, 2019, pp.
  614--619.

\bibitem{aliengo}
\BIBentryALTinterwordspacing
``{Aliengo}.'' [Online]. Available: \url{https://www.unitree.com/en/aliengo}
\BIBentrySTDinterwordspacing

\bibitem{go1}
\BIBentryALTinterwordspacing
``{Go1}.'' [Online]. Available: \url{https://www.unitree.com/en/go1}
\BIBentrySTDinterwordspacing

\bibitem{semini2011design}
C.~Semini, N.~G. Tsagarakis, E.~Guglielmino, M.~Focchi, F.~Cannella, and D.~G.
  Caldwell, ``Design of hyq--a hydraulically and electrically actuated
  quadruped robot,'' \emph{Proceedings of the Institution of Mechanical
  Engineers, Part I: Journal of Systems and Control Engineering}, vol. 225,
  no.~6, pp. 831--849, 2011.

\bibitem{focchi2022clio}
M.~Focchi, M.~Bensaadallah, M.~Frego, A.~Peer, D.~Fontanelli, A.~Del~Prete, and
  L.~Palopoli, ``Clio: a novel robotic solution for exploration and rescue
  missions in hostile mountain environments,'' \emph{arXiv preprint
  arXiv:2209.09693}, 2022.

\bibitem{ur5}
\BIBentryALTinterwordspacing
``{UR5}.'' [Online]. Available:
  \url{https://www.universal-robots.com/products/ur5-robot/}
\BIBentrySTDinterwordspacing

\bibitem{roscia2023orientation}
F.~Roscia, A.~Cumerlotti, A.~Del~Prete, C.~Semini, and M.~Focchi, ``Orientation
  control system: Enhancing aerial maneuvers for quadruped robots,''
  \emph{Sensors}, vol.~23, no.~3, p. 1234, 2023.

\bibitem{ramasubramanian2022digital}
A.~K. Ramasubramanian, R.~Mathew, M.~Kelly, V.~Hargaden, and N.~Papakostas,
  ``Digital twin for human--robot collaboration in manufacturing: Review and
  outlook,'' \emph{Applied Sciences}, vol.~12, no.~10, p. 4811, 2022.

\bibitem{Lieberman2006}
H.~Lieberman, F.~Patern{\`o}, and V.~Wulf, \emph{End user development}.\hskip
  1em plus 0.5em minus 0.4em\relax Springer, 2006, vol.~9.

\bibitem{harris2020array}
\BIBentryALTinterwordspacing
C.~R. Harris, K.~J. Millman, S.~J. van~der Walt, R.~Gommers, P.~Virtanen,
  D.~Cournapeau, E.~Wieser, J.~Taylor, S.~Berg, N.~J. Smith, R.~Kern, M.~Picus,
  S.~Hoyer, M.~H. van Kerkwijk, M.~Brett, A.~Haldane, J.~F. del R{\'{i}}o,
  M.~Wiebe, P.~Peterson, P.~G{\'{e}}rard-Marchant, K.~Sheppard, T.~Reddy,
  W.~Weckesser, H.~Abbasi, C.~Gohlke, and T.~E. Oliphant, ``Array programming
  with {NumPy},'' \emph{Nature}, vol. 585, no. 7825, pp. 357--362, Sep. 2020.
  [Online]. Available: \url{https://doi.org/10.1038/s41586-020-2649-2}
\BIBentrySTDinterwordspacing

\bibitem{joseph2018mastering}
L.~Joseph and J.~Cacace, \emph{Mastering ROS for Robotics Programming: Design,
  build, and simulate complex robots using the Robot Operating System}.\hskip
  1em plus 0.5em minus 0.4em\relax Packt Publishing Ltd, 2018.

\bibitem{frigerio2016robcogen}
M.~Frigerio, J.~Buchli, D.~G. Caldwell, and C.~Semini, ``Robcogen: a code
  generator for efficient kinematics and dynamics of articulated robots, based
  on domain specific languages,'' \emph{Journal of Software Engineering for
  Robotics (JOSER)}, vol.~7, no.~1, pp. 36--54, 2016.

\bibitem{featherstone2014rigid}
R.~Featherstone, \emph{Rigid body dynamics algorithms}.\hskip 1em plus 0.5em
  minus 0.4em\relax Springer, 2014.

\end{thebibliography}
\end{document}